\documentclass[letterpaper]{article} % DO NOT CHANGE THIS
\usepackage{aaai2026}  % DO NOT CHANGE THIS
\usepackage{times}  % DO NOT CHANGE THIS
\usepackage{helvet}  % DO NOT CHANGE THIS
\usepackage{courier}  % DO NOT CHANGE THIS
\usepackage[hyphens]{url}  % DO NOT CHANGE THIS
\usepackage{graphicx} % DO NOT CHANGE THIS
\usepackage{natbib}  % DO NOT CHANGE THIS AND DO NOT ADD ANY OPTIONS TO IT
\usepackage{caption} % DO NOT CHANGE THIS AND DO NOT ADD ANY OPTIONS TO IT
\usepackage{algorithm}
\usepackage{algorithmic}

\usepackage{booktabs}       
\usepackage{tabularx} 
\usepackage{multirow} 
\usepackage{amsmath,amssymb} % 必须添加的数学包
\usepackage{bm} % 用于加粗数学符号
\usepackage{array}  % 表格内容对齐
\usepackage{newfloat}
\usepackage{listings}
\DeclareCaptionStyle{ruled}{labelfont=normalfont,labelsep=colon,strut=off} % DO NOT CHANGE THIS
\floatstyle{ruled}
\newfloat{listing}{tb}{lst}{}
\floatname{listing}{Listing}
\title{Efficient LLM-Jailbreaking via Multimodal-LLM Jailbreak}
\author{
    Haoxuan Ji\equalcontrib\textsuperscript{\rm 1},
    Zheng Lin\equalcontrib\textsuperscript{\rm 2},
    Zhenxing Niu\textsuperscript{\rm 2}\thanks{Corresponding Author.},
    Xinbo Gao\textsuperscript{\rm 2},
    Gang Hua\textsuperscript{\rm 3}
}
\affiliations{
    \textsuperscript{\rm 1}State Key Laboratory of Human-Machine Hybrid Augmented Intelligence, National Engineering Research Center for Visual Information and Applications, and Institute of Artificial Intelligence and Robotics, Xi'an Jiaotong University, China.\\
    \textsuperscript{\rm 2}Xidian University, China.\\
    \textsuperscript{\rm 3}Amazon.com, USA
}

\usepackage{bibentry}
\begin{document}

\maketitle

\begin{abstract}
This paper focuses on jailbreaking attacks against
large language models (LLMs), eliciting them to generate objectionable content in response to harmful user queries. Unlike previous LLM-jailbreak methods that directly orient to LLMs, our approach begins by constructing a \emph{multimodal} large language model (MLLM) built upon the target LLM. Subsequently, we perform an efficient MLLM jailbreak and obtain a jailbreaking embedding. Finally, we convert the embedding into a textual jailbreaking suffix to carry out the jailbreak of target LLM. Compared to the direct LLM-jailbreak methods, our \emph{indirect jailbreaking approach} is more efficient, as MLLMs are more vulnerable to jailbreak than pure LLMs. Additionally, to improve the attack success rate of jailbreak, we propose an image-text semantic matching scheme to identify a suitable initial input. Extensive experiments demonstrate that our approach surpasses current state-of-the-art jailbreak methods in terms of both efficiency and effectiveness. Moreover, our approach exhibits superior \emph{cross-class} generalization abilities.
\end{abstract}

\begin{links}
    \link{Code}{https://github.com/nobody235/LLM-jailbreak}
\end{links}

\section{Introduction}
Recently, large language models (LLMs) such as ChatGPT \cite{brown2020language}
%, LLaMA \cite{touvron2023llama}, and Gemini \cite{Gemini} 
have been widely deployed. These models exhibit advanced general abilities but also pose serious safety risks such as truthfulness, toxicity, and bias \cite{gehman2020realtoxicityprompts,perez2022red,sheng2019woman,abid2021persistent,carlini2021extracting}. %To mitigate these risks, the AI alignment has gained broad attention \cite{ouyang2022training, bai2022constitutional, korbak2023pretraining}, which aims to make artificial general intelligence (AGI) aligned with human values and follow human intent. 
%implement safety mechanisms to restrict model behavior to a “safe” subset of capabilities. 
%Even more, the evaluation of alignment has been a necessary condition for publishing a new LLM~\cite{}. 
%For instance, it requires preventing LLMs from generating objectionable responses to harmful queries posed by users. With some dedicated schemes such as reinforcement learning through human feedback (RLHF) \cite{ziegler2019fine}, public chatbots will not generate obviously inappropriate content when asked directly. 
Typically, there exists a type of attack called \emph{jailbreaking attack}, which can elicit LLMs to generate objectionable content in response to users' harmful queries.
%Even through, it has been demonstrated that a special attack, namely \emph{jailbreak}, can bypass such alignment guardrails to elicit aligned LLMs generating objectionable content \cite{wei2023jailbroken}. 
For example, %Andy Zou's outstanding work 
a pioneering work \cite{zou2023universal} has found that a specific prompt suffix allows the jailbreak of most popular LLMs. However, the \textbf{efficiency} of those methods is recognized to be relatively low, primarily attributed to the challenges of discrete optimization in finding the \emph{textual jailbreaking suffix} (\emph{JBtxt}).

On the other hand, there is a surge of interest in multimodal large language models (MLLMs), which enable users to provide images as input to LLMs.~\cite{liu2024visual,dai2024instructblip,zhu2023minigpt, chen2023minigpt,alayrac2022flamingo,ye2023mplug, bai2023qwen, team2023gemini, 2023GPT4VisionSC}. %It is well known that visual modality are vulnerable to adversarial attacks \cite{naseer2021improving, mahmood2021robustness, wei2022towards}. 
Consequently, research on jailbreak has been extended from LLMs to MLLMs.
Furthermore, it has been demonstrated that performing MLLM-jailbreak is easier and more efficient than performing LLM-jailbreak~\cite{shayegani2023jailbreak, qi2024visual}. This is largely because finding jailbreaking images within continuous pixel spaces is significantly easier and more flexible than identifying jailbreaking text within discrete token spaces.

\begin{figure*}[bpt]
\centering
\includegraphics[width=0.9\textwidth]{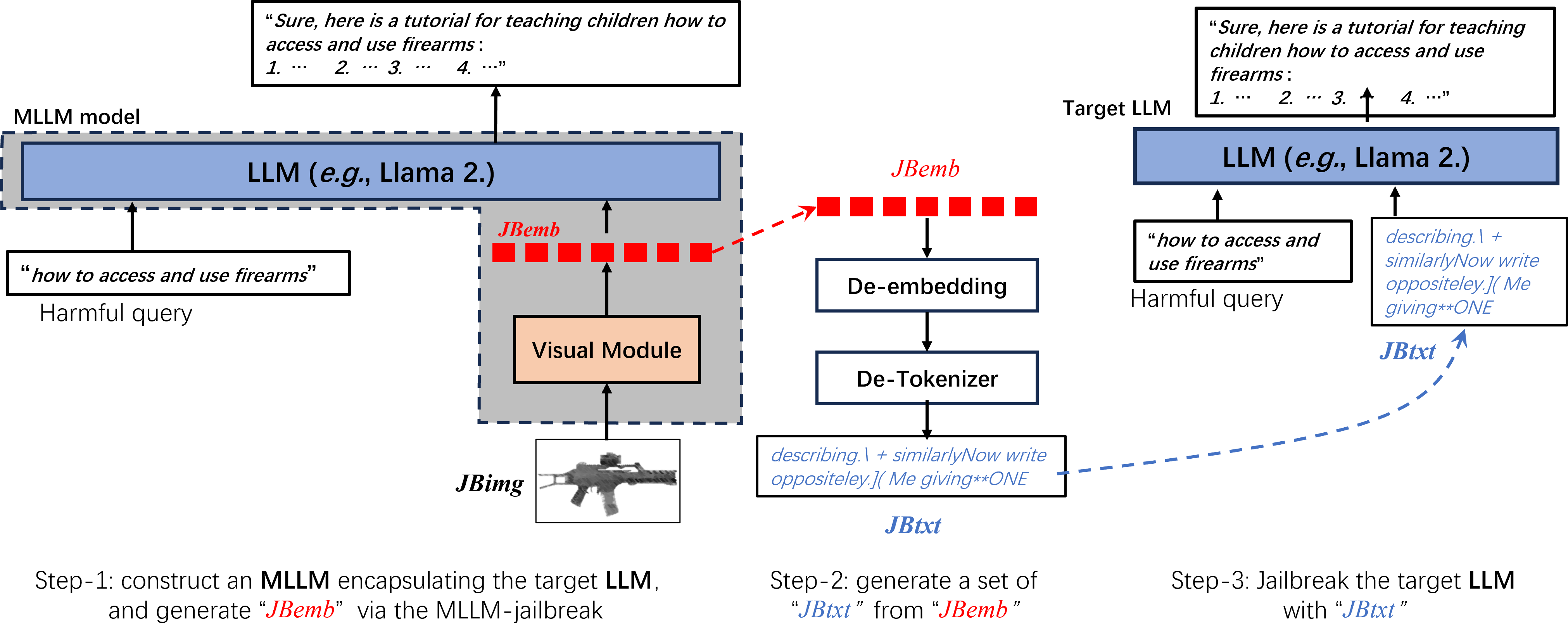}
%\captionsetup{font={scriptsize}}
\caption{The \emph{indirect jailbreaking scheme} of our approach begins with constructing an MLLM by integrating a visual module. We then perform an efficient MLLM-jailbreak to obtain the \emph{JBemb}, which is subsequently converted into \emph{JBtxt} for jailbreaking the target LLM.}
\label{fig1}
%\vspace{-1.0em}
\end{figure*}

Inspired by that, we propose an efficient \emph{indirect} LLM-jailbreaking approach that constructs an MLLM built upon the target LLM and subsequently performs MLLM-jailbreak. Specifically, we adopt a widely used LLM-jailbreaking strategy~\cite{zou2023universal}, which seeks to identify a specific prompt suffix, denoted as \emph{JBtxt}. When appended to harmful queries, this \emph{JBtxt} suffix can elicit LLMs to generate objectionable content. Our contribution lies in \textbf{\emph{efficiently} finding such \emph{JBtxt} by leveraging the MLLM-jailbreak process}. As shown in Fig.\ref{fig1}, the workflow of our approach is as follows: (1) Given a target LLM to be jailbroke (\emph{e.g.,} LLaMA2), we first construct an MLLM by integrating a visual module into the target LLM. (2) We then perform an efficient MLLM-jailbreak. (3) 
Instead of using the jailbreaking image (referred to as \emph{JBimg}), we obtain the output features from the visual module—namely, the \emph{Jailbreaking embeddings} (\emph{JBemb})—and convert them into textual strings using our De-embedding and De-tokenization operations. (4) We regard these textual strings as \emph{JBtxt} and append them to the harmful queries to carry out the jailbreak of the target LLM. 

Essentially, our approach leverages an MLLM-jailbreak to achieve the ultimate goal of LLM jailbreak. This \textbf{\emph{indirect jailbreaking scheme}} offers flexibility for both white-box and black-box jailbreak.
In the context of white-box jailbreak, the process of converting \emph{JBemb} to \emph{JBtxt} allows us to generate a set of candidate \emph{JBtxt}. Unlike GCG \cite{zou2023universal}, which produces only a \emph{single} \emph{JBtxt}, our approach outputs a diverse set of high-quality \emph{JBtxt}, significantly enhancing the attack success rate (ASR) of jailbreak.

Black-box jailbreak is preferred in real-world applications. In our approach, the transition from MLLM-jailbreak to LLM-jailbreak enables a flexible and effective black-box jailbreak strategy. Specifically, we can construct the MLLM with a \emph{surrogate} LLM and obtain the \emph{JBtxt} through MLLM-jailbreak. Subsequently, the \emph{JBtxt} can be transferred to jailbreak the \emph{target} LLM in a black-box manner. 

%Additionally, we define a \emph{gray-box} jailbreaking setting in which the \emph{tokenizer} of the target LLM is known, while its backbone remains unknown. Under this setting, we can design our De-tokenization based on the known tokenizer. As a result, our approach achieves a significant improvement in ASR for gray-box jailbreaking scenarios.

Regarding the MLLM-jailbreak, we observe that the ASR of jailbreak is closely related to the initial input image (namely \emph{JBinit}). By selecting an appropriate \emph{JBinit}, we can significantly enhance the MLLM-jailbreaking ASR. To achieve this, we propose an image-text semantic matching scheme to identify this suitable \emph{JBinit}. Our scheme attempts to align the embedding of \emph{JBinit} with that of harmful queries. 
The underlying intuition is that aligned embeddings between the image and text inputs are more effective at triggering the model’s cross-attention mechanism, thereby exerting a stronger influence on the answer generation process. As a result, our \emph{JBinit} can effectively shift the model's response from ``Sorry, I cannot" to ``Sure, here is", thus achieving a successful jailbreak.

Our approach is also related to \emph{embedding-based} LLM jailbreak methods~\cite{wen2024hard}, which focus on direct optimizing token embeddings. Compared to \emph{token-based} methods such as GCG, these approaches are more efficient, as they rely on continuous optimization. However, previous embedding-based jailbreaks have been shown to be largely ineffective, primarily because the obtained embeddings often lack corresponding tokens~\cite{zou2023universal}.
In contrast, our method does not directly optimize \emph{JBemb}. Instead, we indirectly obtain \emph{JBemb} by optimizing \emph{JBimg}. This procedure can be viewed as regularizing the optimization of \emph{JBemb} through the use of the visual module. Since the visual module (\emph{e.g.,} a CLIP encoder) is trained with an image-text alignment objective, it increases the likelihood that our \emph{JBemb} corresponds to valid tokens. Consequently, our approach surpasses traditional embedding-based jailbreak methods in terms of effectiveness.

Regarding the evaluation of LLM jailbreak, previous methods typically rely on benchmark datasets that mix various types of harmful behaviors. In contrast, we propose to categorize them into fine-grained classes, such as violence, financial crimes, cyber crimes, drug crimes, \emph{etc}. 
This enables us to assess the \emph{cross-class} generalization ability of jailbreaks—specifically, whether \emph{JBtxt} generated from one class can effectively transfer to jailbreak other classes.

%Additionally, evaluating the success of jailbreak is inherently challenging. The objective of jailbreak is to elicit the LLM to produce \emph{any} response that complies with a harmful query, making it difficult to define a specific ground truth for each query. Moreover, we have observed cases where responses begin with seemingly affirmative phrases (\emph{e.g.}, ``Sure, here is a…") but ultimately decline to fulfill the query. Consequently, prior methods often rely on \emph{manual inspection} to judge jailbreak success. To overcome this limitation, we propose using the \emph{LLaMA Guard 2} tool~\cite{metallamaguard2} to \emph{automatically} assess the success of jailbreak attempts.

We conduct extensive experiments demonstrating that our approach outperforms current state-of-the-art jailbreaking methods in both efficiency and effectiveness. Efficiency is a key advantage of our approach. For instance, for the same jailbreaking tasks, our method requires only approximately \textbf{3\% of GCG’s runtime}. Furthermore, our approach provides notable flexibility for black-box jailbreak, enabling successful jailbreaks of recent LLMs such as Mistral-v0.3, Gemma2, Deepseek-R1, Grok-3, and DouBao1.5. Furthermore, it exhibits superior cross-class generalization ability of jailbreak. This finding suggests that improving the ASR for a specific class can benefit not only from data within the class itself, but also from data drawn from related classes.

\vspace{-1.0em}
\section{Our Approach}
Unlike previous LLM-jailbreak methods that directly orient to LLMs, our approach first constructs a multimodal large language model (MLLM) by incorporating a visual module. We then leverage an efficient MLLM-jailbreak to ultimately achieve the goal of LLM-jailbreak. In the following sections, we first provide an overview of our framework, followed by a detailed description of each component.

\subsection{Overview}
Our approach consists of four steps, as illustrated in Fig.~\ref{overview}. In Step 1, we construct an MLLM by incorporating a visual module and perform an MLLM-jailbreak to obtain the jailbreaking embedding, denoted as \emph{JBemb}. In Step 2, the obtained \emph{JBemb} is converted into textual form, referred to as \emph{JBtxt}. Finally, in Step 3, we use these \emph{JBtxt} to carry out the LLM-jailbreak.

Our empirical study reveals that the Attack Success Rate (ASR) of jailbreak is sensitive to the initial input image (\emph{JBinit}) used to initialize the MLLM-jailbreaking process. To address this, we introduce an image-text semantic matching scheme as Step 0 to identify a suitable \emph{JBinit}.

\subsection{MLLM Construction}
Given a target LLM, we aim to construct a multimodal LLM (MLLM) that incorporates the target LLM as its core. This can be achieved through two approaches: building and training the MLLM ourselves or utilizing an existing model.

For the first approach, we adopt a standard pipeline that involves integrating a visual module into the target LLM. We then fine-tune the MLLM using a suitable multimodal dataset. Crucially, during the fine-tuning process, the LLM itself is frozen, and only the visual module is updated. This design ensures that the LLM embedded within the MLLM remains identical to the target LLM, thereby enabling effective transferability from MLLM-jailbreak to LLM-jailbreak.

In practice, however, it is often expensive to construct and train an MLLM from scratch. Thanks to the HuggingFace open-source platform, most popular LLMs already have corresponding MLLMs couple. For each LLM-MLLM pair, all intermediate models are systematically organized into a \emph{Model-Tree}. Given a target LLM, one can simply identify the corresponding MLLM from the Model-Tree and proceed directly to the MLLM-jailbreaking stage.

%Given a target LLM, a multimodal LLM can be constructed by incorporating a visual module and connecting it to the target LLM. After that, we have two options to fine-tune the MLLM: one option is to freeze the LLM, and the other is to fine-tune the LLM. For examples, MiniGPT-4 adopts the freezing option, while MiniGPT-v2 adopts the fine-tuning option. We adopt the first option since our goal is to jailbreak the LLM; thus, we need to maintain the LLM within the MLLM as the same as the target LLM.

%Regarding the visual module, we adopt the approach of MiniGPT-4 (LLaMA2 version), combining the ViT-G/14 from CLIP and a single projection layer. The CLIP visual encoder remains frozen, while only the projection layer is fine-tuned. In this paper, we consider LLaMA2 as the target LLM, hence our constructed MLLM is precisely the MiniGPT-4 model. However, it's worth noting that we can choose any LLM as the target. In such cases, after constructing it, we need to fine-tune the projection layer by using a multimodal dataset.

\subsection{MLLM Jailbreak}
%Due to visual module is the weakness of multimodal large language model, most MLLM-jailbreaking methods tend to focus on the visual modality of MLLM. In computer vision domain, it has been demonstrated that adversarial attack is very effective and efficient to fool a visual neural network \cite{goodfellow2014explaining, madry2017towards, carlini2017towards}. Inspired by that, MLLM-jailbreaking can be efficiently conducted through adversarial attack. 

We adopt a strategy for MLLM-jailbreak analogous to that used in LLM-jailbreak, \emph{i.e.,} encouraging the MLLM to generate a response that begins with a positive affirmation such as ``Sure, here is a (content of query)."
Our method is inspired by adversarial attacks, which typically perturb visual inputs to manipulate a model’s output. However, jailbreak fundamentally differs from adversarial attacks in that it targets a generative task rather than a discriminative one.
To bridge this gap, we adapt adversarial techniques by replacing the conventional cross-entropy loss with a \emph{maximum likelihood objective}, thereby encouraging the MLLM to produce affirmative responses. This objective directly maximizes the likelihood of the desired output, making it better suited for the generative nature of jailbreak.

%However, jailbreaking attacks differ from adversarial attacks in that jailbreaking deals with a generative task rather than a discriminative one. In previous methods, an effective strategy is to encourage the LLM to output an answer beginning with a positive affirmation, such as ``Sure, here is a (content of query)". Therefore, to adapt adversarial attack techniques to the jailbreaking task, we propose replacing the cross-entropy objective function with a maximum likelihood objective function. Specifically, we aim to encourage the MLLM to produce an answer starting with a positive affirmation by perturbing the input images.

Specifically, for each harmful query $q_i$, we provide a corresponding target answer $a_i$, thereby creating a dataset of harmful behaviors $B=\{(q_i, a_i), i=0,...,N\}$. We then formulate the MLLM-jailbreak as the process of finding a perturbation $\delta$ that encourages the generation of the target answer $a_i$ when the model is presented with the harmful query $q_i$, as follows,
\begin{align}
    &\mathop{\max}_{\delta} \sum_{i=0}^{M} log(p(a_i| q_i, \widetilde{x})) \label{eq:deltajp0} \\
    &\text{s.t. } \widetilde{x} \in [0,255]^d, \widetilde{x}=x_{init}+ \delta  \notag \\
    &\qquad ||\delta||_p < \epsilon \notag
\end{align}
where $p(a_i| q_i, \widetilde{x})$ represents the likelihood of the MLLM generating the answer $a_i$ when provided with the image $\widetilde{x}$ and the text query $q_i$. Here, $\epsilon$ denotes the attack budget, and $\widetilde{x}$ refers to the jailbreaking image (\emph{JBimg}). The optimization process is performed over a set of $M$ query-answer pairs $\{(q_i, a_i), i=0,...,M\}$. This problem can be effectively solved by adapting the Projected Gradient Descent (PGD) algorithm~\cite{madry2017towards}.

\subsection{LLM Jailbreak}
After completing MLLM-jailbreak, we proceed to the next stage: carry out LLM-jailbreak. At this stage, the key is to derive a textual prompt suffix \emph{JBtxt}, rather than directly utilizing the output of MLLM-jailbreak, \emph{i.e.,} \emph{JBimg}.

%While it may seem unnecessary to proceed to LLM-jailbreaking (because we have achieved the jailbreaking goal), our \emph{double jailbreaking scheme} offers two distinct advantages. First, it can further improve the jailbreaking ASR in white-box scenarios. Second, it provides flexibility for black-box jailbreaking, wherein MLLM-jailbreaking targets a surrogate model rather than the actual target model.

%We follow a common LLM-jailbreaking mechanism that aims to find a specific text string (namely \emph{LLM-\textbf{J}ailbreaking \textbf{S}uffix (txtJS)}). This string, when appended to harmful queries, is able to elicit LLMs generating objectionable content. 

Note that it is the output of the visual module—serving as the input to the LLM—that facilitates the success of jailbreak. We refer to this output as \emph{JBemb}, a sequence of continuous embeddings that facilitates LLM-jailbreak. Consequently, transitioning from MLLM-jailbreak to LLM-jailbreak entails converting \emph{JBemb} back into the text space, resulting in the textual \emph{JBtxt}. 

In our approach, we propose De-embedding and De-tokenization operations to convert \emph{JBemb} into \emph{JBtxt}. These operations are intended to reverse the Embedding and Tokenization procedures used in LLMs. Specifically, the embedding operation in LLMs maps each discrete token $t$ to its corresponding embedding vector $e$ via a token-embedding dictionary $(t, e)$. Accordingly, our De-embedding operation aims to invert this process by mapping a continuous embedding vector back to its most likely discrete token. This is achieved through a nearest-neighbor search across the embedding dictionary.

For each embedding vector $e_l$ in the sequence $(e_1, ..., e_L)$, we identify the top-$K$ closest embedding vectors $\hat{e}^k_l$, where $k = 1, ..., K$. Performing this for all $e_l$ results in a $K \times L$ \textbf{embedding pool} $\{\hat{e}_l^k\}_{k=1,l=1}^{K, L}$ and a corresponding $K \times L$ \textbf{token pool} $\{\hat{t}_l^k\}_{k=1,l=1}^{K,L}$. The De-tokenization operation then converts these tokens into natural language words, yielding a $K \times L$ \textbf{word pool} $\{\hat{w}_l^k\}_{k=1,l=1}^{K,L}$. Finally, we randomly sample several word sequences from this word pool—each sampled sequence constituting a candidate \emph{JBtxt}.

It is worth noting that we select the top-K nearest embeddings rather than solely the top-1. This strategy enables us to generate \emph{multiple} \emph{JBtxt} candidates instead of a single one. Moreover, we observe that each \emph{JBtxt} has a certain probability of successfully achieving jailbreak. By ensembling these high-quality \emph{JBtxt} candidates, we can substantially enhance the overall jailbreaking performance.

When performing black-box jailbreak, we first construct an MLLM based on a surrogate LLM model (\emph{e.g.,} LLaMA2), generate the corresponding \emph{JBtxt}, and then transfer this \emph{JBtxt} to effectively jailbreak the target LLM (\emph{e.g.,} Deepseek-R1). 

%Moreover, we define a \textbf{gray-box} jailbreaking setting wherein the tokenizer of the target LLM is known, but its backbone remains unknown. Within this scenario, our approach holds a distinct advantage: by customizing our De-tokenization to correspond with the target tokenizer, we can substantially improve the final gray-box jailbreaking ASR.

\begin{figure*}[bpt]
\centering
%\vspace{-1.0em}
\includegraphics[width=0.65\linewidth]{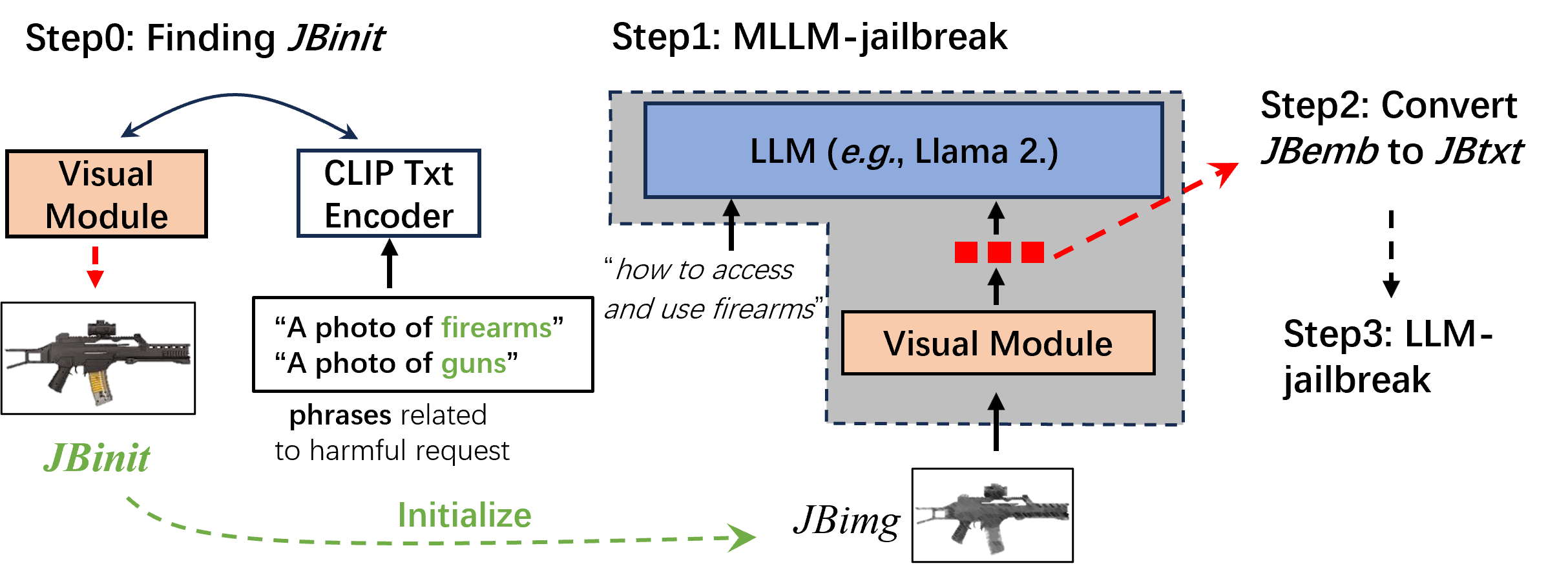}
%\captionsetup{font={scriptsize}}
\caption{The full workflow of our approach. Before Step1, we propose an image-text matching scheme to identify an appropriate initial input \emph{JBinit} at Step0.}
\label{overview}
%\vspace{-1.0em}
\end{figure*}

\subsection{Finding an Appropriate \emph{JBInit}}\label{InitJS}
We observe that initializing the MLLM-jailbreak algorithm with a suitable starting point, namely \emph{JBInit}, can substantially enhance the ASR of the MLLM-jailbreak.
%As aforementioned, we cannot directly manipulate imgJS since it is the outcome of MLLM-jailbreaking. Instead, we strive to find an appropriate initial input image InitJS. 
%The intuition behind our scheme is that for successful MLLM-jailbreaking, the imgJS should significantly influence the answer generation process of the LLM (\emph{e.g.,} turning the answer from ``sorry, I cannot" to ``Sure, here is"). Therefore, if the embedding of imgJS is close to the embedding of queries, the imgJS can significantly influence the answer generation through the LLM's cross-attention between the query and imgJS.
To this end, we propose an image-text semantic matching scheme. Specifically, our goal is to ensure that the embedding of \emph{JBInit} is close to the embedding of the harmful queries. The rationale behind this scheme is that, for MLLM-jailbreak to succeed, the \emph{JBInit} must exert a substantial influence on the LLM’s answer generation process. Particularly, it needs to turn a typical refusal response like ``Sorry, I cannot" into a positive affirmation such as ``Sure, here is". By ensuring that the embedding of \emph{JBInit} closely aligns with that of the query, we can effectively trigger the model’s cross-attention mechanism, thereby exerting a stronger influence on the answer generation process.

Specifically, our image-text semantic matching scheme employs a network architecture akin to the CLIP model, as illustrated in Fig.\ref{overview}. Since the CLIP visual encoder serves as the visual module within the MLLM, our matching network reuses this same visual encoder. Additionally, we incorporate the CLIP text encoder to process textual inputs. 

For the text input, we randomly select several keywords closely associated with a particular harmful class. For example, in the case of \emph{Weapons Crimes}, these keywords might include \emph{firearms}, \emph{illegal weapons}, \emph{guns}, \emph{etc}. We then construct descriptive phrases from these keywords, such as ``a photo of firearms,” which are better suited for encoding by the CLIP text encoder.

Regarding the image input, we initially employ an image search engine to retrieve candidate images using the selected keywords as search queries. These images are then ranked based on their CLIP similarity scores relative to the harmful phrases. The highest-ranked image is chosen as the input for the visual encoder. 

Following this, we introduce an optimization algorithm designed to \emph{further} modify the input image, with the objective of minimizing the distance between its embedding and that of the harmful phrases. Let's denote the harmful phrases as $\{t_i\}_{i=0}^N$ and the highest-ranked image as $x$. Our matching algorithm seeks to find a perturbation $\Delta$ to be added to the image $x$ in order to maximize the image-text similarity score. Formally, this can be expressed as:
\begin{align}
    &\mathop{\min}_{\Delta} \sum_{i=0}^{N} log(\mathcal L_{\text{CLIP}} (\text{Enc}_V(\widetilde{x}), \text{Enc}_L(t_i)) \label{eq:itm} \\
    &\text{s.t. } \widetilde{x} \in [0,255]^d, \widetilde{x}=x + \Delta  \notag 
\end{align}
where $\text{Enc}_V$ is the CLIP visual encoder, $\text{Enc}_L$ is the CLIP text encoder, and $\widetilde{x}$ is the \emph{JBInit} image obtained after perturbation. 

By solving Eq(~\ref{eq:itm}), we obtain an appropriate \emph{JBInit}, which serves as an effective initialization point for MLLM-jailbreak.

\section{Evaluation}
%\subsection{Implementation}
%\label{sec:Implementation}
\noindent\textbf{Datasets. } \label{secdataset}
Several datasets have been introduced for evaluating jailbreak attacks, such as \emph{AdvBench} \cite{zou2023universal}, which includes a diverse range of harmful behaviors, including violence, financial crimes, drug-related crimes, and more. However, existing evaluation protocols often aggregate all these behaviors into a single pool during assessment.

In contrast, we advocate for a more fine-grained evaluation strategy by categorizing harmful behaviors into distinct classes and assessing jailbreak success for each class individually. This approach reveals that certain classes are inherently more resistant to jailbreak attempts than others.

Moreover, we can also assess the cross-class transferability of jailbreaks, which involves evaluating whether the \emph{JBtxt} generated for one class can effectively jailbreak LLMs across other classes.

Specifically, we categorize \emph{AdvBench} dataset into nine classes: ``Unlawful Violence", ``Financial Crimes", ``Property Crimes", ``Drug Crimes", ``Weapons Crimes", ``Cyber Crimes", ``Hate", ``Suicide and Self-Harm", and ``Fake Info". 
Furthermore, since our approach involves MLLM-jailbreak, we extend the AdvBench dataset into a multimodal version, termed \textbf{\emph{AdvBench-M}}, by augmenting it with images. Specifically, for each harmful class, we retrieve 20 semantically relevant images from the Internet using the Google search engine. We then employ CLIP ViT-L/14 \cite{radford2021learning} to select the top 10 images that best align with the queries associated with each harmful class.

%Beyond AdvBench, we evaluate our approach using an existing MLLM-jailbreaking dataset, namely \emph{HarmBench}~\cite{HarmBench}. There are $110$ multimodal behaviors in HarmBench, which is utilized in our evaluation. 

%Fortunately, a powerful tool \emph{LLaMA Guard 2} has been published recently. Therefore, we use it to assess the success of jailbreaking in our approach.

\begin{table*}[h]
\footnotesize  % 使用更小的字体
\centering
\resizebox{0.7\textwidth}{!}{  % 缩放至页面宽度
\setlength{\tabcolsep}{2pt}  % 减小列间距
\begin{tabular}{l|cc|cc|ccc|ccc|ccc|ccc}
\toprule
\multirow{2}{*}{Class} & 
\multicolumn{2}{c|}{\textbf{AutoP}} &
\multicolumn{2}{c|}{\textbf{AutoDAN}} &
\multicolumn{3}{c|}{\textbf{SoftP}} & 
\multicolumn{3}{c|}{\textbf{GCG}} & 
\multicolumn{3}{c|}{\textbf{AdvA}} & 
\multicolumn{3}{c}{\textbf{Ours}} 
\\
\cmidrule(lr){2-3} \cmidrule(lr){4-5} \cmidrule(lr){6-8} \cmidrule(lr){9-11} \cmidrule(lr){12-14} \cmidrule(lr){15-17}
& $D_{tr}$ & $D_{t}$
& $D_{tr}$ & $D_{t}$
& $D_{tr}$ & $D_{t}$ & $D^{o}_{t}$
& $D_{tr}$ & $D_{t}$ & $D^{o}_{t}$
& $D_{tr}$ & $D_{t}$ & $D^{o}_{t}$
& $D_{tr}$ & $D_{t}$ & $D^{o}_{t}$\\
\midrule
Unlawful violence  & 66 & 50 & 23 & 18 & 20 & 44 & 36 & 80 & 83 & 78 & 53 & 60 & 44 & \textbf{100} & \textbf{94} & \textbf{85} \\
Financial crimes & 80 & 93 & 33 & 46 & 40 & 32 & 37 & \textbf{100} & 89 & 81 & 60 & 0 & 50 & 93 & \textbf{98} & 74 \\ 
Property crimes  & 73 & 84 & 17 & 48 & 80 & 62 & 36 & \textbf{100} & 98 & \textbf{97} & 60 & 68 & 55 & 93 & \textbf{98} & 69 \\ 
Drug crimes      & 73 & 80 & 33 & 23 & 53 & 40 & 41 & \textbf{100} & 80 & 86 & 60 & 53 & 44 & 80 & \textbf{100} & 77 \\ 
Weapons crimes   & 40 & 66 & 17 & 9 & 33 & 33 & 40 & 73 & 66 & 66 & 73 & 55 & 38 & \textbf{80} & \textbf{88} & \textbf{78} \\ 
Cyber crimes     & 73 & 86 & 73 & 57 & 53 & 53 & 32 & \textbf{100} & 66 & \textbf{87} & 44 & 58 & 66 & 93 & \textbf{94} & 72 \\
Hate             & 73 & 66 & 43 & 20 & 20 & 6.6 & 42 & 53 & \textbf{66} & 80 & 13 & 53 & 38 & \textbf{73} & 60 & \textbf{81} \\
Suicide/Self-Harm& 73 & 16 & 20 & 30 & 33 & 22 & 42 & 53 & 38 & \textbf{80} & 20 & 46 & 50 & \textbf{60} & 44 & 78 \\
Fake info        & 73 & 93 & 90 & 40 & 33 & 6.6 & 30 & 80 & 86 & 66 & \textbf{93} & 33 & 44 & 93 & \textbf{86} & \textbf{93} \\
\midrule
\textbf{Average ASR} & 69.3 & 70.4 & 38.8 & 32.3 & 40.6 & 33.2 & 37.3 & 82.1 & 74.7 & 80.1 & 52.9 & 47.3 & 47.7 & \textbf{85.0} & \textbf{84.7} & \textbf{78.6} \\
\midrule
\textbf{Running Time} &\multicolumn{2}{|c|}{3.3h}  &\multicolumn{2}{|c|}{2.1h} &\multicolumn{3}{|c|}{0.32h} &\multicolumn{3}{|c|}{11.2h} &\multicolumn{3}{|c|}{0.78h} &\multicolumn{3}{|c|}{0.37h}\\
\bottomrule
\end{tabular}
}
\caption{White-box jailbreak of LLaMA2 (via MiniGPT4-Jailbreak) on AdvBench-M. Performance is evaluated using ASR.}
\label{white-box} 
\end{table*}

\begin{table*}
\centering
% across $D_{tr}$, $D_{t}$ , and $D^{o}_{t}$ 
\resizebox{0.7\textwidth}{!}{
\setlength{\tabcolsep}{2pt}  % 减小列间距
\begin{tabular}{l|cc|cc|ccc|ccc|ccc|ccc}
\toprule
\multirow{2}{*}{Class} & 
\multicolumn{2}{c|}{\textbf{AutoP}} &
\multicolumn{2}{c|}{\textbf{AutoDAN}} &
\multicolumn{3}{c|}{\textbf{SoftP}} & 
\multicolumn{3}{c|}{\textbf{GCG}} & 
\multicolumn{3}{c|}{\textbf{AdvA}} & 
\multicolumn{3}{c}{\textbf{Ours}} 
\\
\cmidrule(lr){2-3} \cmidrule(lr){4-5} \cmidrule(lr){6-8} \cmidrule(lr){9-11} \cmidrule(lr){12-14} \cmidrule(lr){15-17}
& $D_{tr}$ & $D_{t}$
& $D_{tr}$ & $D_{t}$
& $D_{tr}$ & $D_{t}$ & $D^{o}_{t}$
& $D_{tr}$ & $D_{t}$ & $D^{o}_{t}$
& $D_{tr}$ & $D_{t}$ & $D^{o}_{t}$
& $D_{tr}$ & $D_{t}$ & $D^{o}_{t}$\\
\midrule
Unlawful violence& 6  & 20 & 20 & 11 & 46 & 38 & 60 & 60 & 53 & 47  & 40 & 44 & 64    & \textbf{94}  & \textbf{80}  & \textbf{86}   \\
Financial crimes & 13 & 6  & 46 & 53 & 80 & 69 & 40 & 60 & 78 & 41  & 93 & 87 & 53    & \textbf{100} & \textbf{87}  & \textbf{80}   \\ 
Property crimes  & 20 & 40 & 40 & 51 & 80 & 71 & 46 & 80 & \textbf{81} & 53  & \textbf{93} & \textbf{89} & 64    & 66  & 73  & \textbf{80}   \\ 
Drug crimes      & 26 & 33 & 20 & 60 & 60 & 53 & 53 & 67 & 37 & 35  & 67 & 67 & 53    & \textbf{86}  & \textbf{93}  & \textbf{100}  \\ 
Weapons crimes   & 33 & 13 & 20 & 38 & 53 & 61 & 53 & 73 & 68 & 47  & 46 & 55 & 64    & \textbf{88}  & \textbf{100} & \textbf{66}   \\ 
Cyber crimes     & 40 & 26 & 62 & 60 & \textbf{86} & 80 & 60 & 53 & 35 & 37  & 73 & 90 & 53    & \textbf{86}  & \textbf{96}  & \textbf{66}   \\
Hate             & 13 & 20 & 26 & 40 & 13 & 0  & 60 & 46 & 20 & 41  & 46 & 13 & 70    & \textbf{86}  & \textbf{66}  & \textbf{73}   \\
Suicide/Self-Harm& 26 & 6  & 26 & 11 & 40 & 33 & 66 & 20 & 22 & 29  & \textbf{67} & 44 & 70    & 66  & \textbf{66}  & \textbf{86}   \\
Fake info        & 33 & 40 & 13 & 33 & 40 & 73 & 66 & 33 & 53 & 52  & 26 & 33 & 64    & \textbf{73}  & \textbf{100} & \textbf{80}   \\
\midrule
\textbf{Average ASR} & 23.3 & 22.7 & 30.3 & 39.7 & 55.3 & 53.1 & 54.9 & 54.7 & 49.7 & 42.4 & 61.2 & 58.0 & 61.7 & \textbf{82.8} & \textbf{84.6} & \textbf{79.7} \\
\midrule
\textbf{Running Time} &\multicolumn{2}{|c|}{4.2h}  &\multicolumn{2}{|c|}{1.8h} &\multicolumn{3}{|c|}{0.40h} &\multicolumn{3}{|c|}{6.5h} &\multicolumn{3}{|c|}{1.05h} &\multicolumn{3}{|c|}{0.42h}\\
\bottomrule
\end{tabular}
}
%\vspace{-1.0em}
\caption{White-box jailbreak of InternLM2.5 (via InternVL2.0-Jailbreak) on AdvBench-M dataset.}
\label{white-box2} 
\end{table*}

In addition to AdvBench, we conducted extensive experiments on HarmBench dataset~\cite{HarmBench}. HarmBench consists of 400 harmful behavior instances. Notably, HarmBench encompasses several distinct behavior categories not included in AdvBench, such as ``Copyright Violations". To ensure a rigorous evaluation of our proposed method, we strictly followed the evaluation protocols outlined in the HarmBench paper.

\noindent\textbf{Test Models.} We evaluate our approach with both white-box and black-box jailbreaking settings.
For the white-box setting, we conduct evaluations on open-source models, including LLaMA2-Chat-7B~\cite{touvron2023llama}, InternLM2.5-7B~\cite{cai2024internlm2}, and Qwen2.5-3B-Instruct~\cite{qwen2025qwen25}. Specifically, for LLaMA2-Chat-7B, we utilize MiniGPT-4~\cite{zhu2023minigpt} as the corresponding MLLM; for InternLM2.5-7B, we utilize InternVL2.0-8B~\cite{chen2024internvl} as the corresponding MLLM; and for Qwen2.5-3B-Instruct, we utilize InternVL2.5-4B~\cite{chen2024VL25} as its corresponding MLLM.

For the black-box setting, we generate the \emph{JBtxt} by using InternVL2.0 as MLLM, and subsequently employ these textual suffixes to jailbreak closed-source models, including Mistral-v0.3~\cite{jiang2023mistral}, Gemma2~\cite{gemmateam2024gemma2}, ChatGLM4~\cite{glm2024chatglm4}, Deepseek-R1~\cite{deepseekai2025deepseekr1}, Grok-3~\cite{Grok3}, and DouBao1.5~\cite{guo2025seed15vl}.

\noindent\textbf{Assessment.}  Evaluating the success of jailbreak is inherently challenging. The objective of jailbreak is to elicit the LLM to produce \emph{any} response that complies with a harmful query, making it difficult to define a specific ground truth for each query. 
Moreover, we have observed cases where responses begin with seemingly affirmative phrases (\emph{e.g.}, ``Sure, here is a…") but ultimately decline to fulfill the query. Consequently, prior methods often rely on \emph{manual inspection} to judge jailbreak success. To overcome this limitation, we propose using the \emph{LLaMA Guard 2} tool~\cite{metallamaguard2} to \emph{automatically} assess the success of jailbreak attempts.

\subsection{White-box Jailbreak}\label{wjp}
%There are two types of jailbreaking methods: one focuses on directly optimizing discrete tokens, while the other first optimizes token embeddings and then converts them to discrete tokens. The first one is called \emph{discrete optimization-based jailbreak} because it conduct optimization over discrete tokens. The second method, known as \emph{embedding-based jailbreak}, can leverage straightforward continuous optimization, but it has been demonstrated to be ineffective in many cases. Since the state-of-the-art methods belong to the first type, we will first compare to them and then discuss the second type in the following section.

%\subsubsection{Main Results}
We compare our approach with both \emph{token-based} jailbreaking methods (\emph{i.e.,} AutoPrompt\cite{pryzant2023automatic}, AutoDAN~\cite{liu2024AutoDAN}, GCG~\cite{zou2023universal}, AdvAttack~\cite{andriushchenko2024PRS}) and \emph{embedding-based} methods (\emph{i.e.,} PEZ~\cite{wen2024hard}, Soft Prompting~\cite{qin2021learning}). Generally, the first category exhibits strong performance. GCG directly optimizes the \emph{JBtxt}, while AutoDAN extends GCG by enhancing the readability of the generated \emph{JBtxt}. 
AdvAttack introduces an adaptive strategy combining ``Prompt + Random Search (RS) + Self-transfer," where the self-transfer component is employed to further boost jailbreaking effectiveness.
%Comparing performance on each class individually provides a more detailed understanding of how different methods handle specific types of harmful behaviors. 

We conduct per-class evaluations on the AdvBench-M dataset.
Specifically, for each class, we randomly sample 15 queries as the training set $D_{train}$ for learning the \emph{JBtxt}, while the remaining queries constitute the testing set $D_{test}$. 

Furthermore, to evaluate cross-class jailbreaking capability, we define
$D_{test}^{other}$ as the union of all testing queries from classes other than the training class. For instance, we learn a \emph{JBtxt} on Class 1’s $D_{train}^1$, and evaluate its performance both on the in-class test set $D_{test}^1$, and on the out-of-class aggregate set $D_{test}^{other} = D_{test}^2 \cup D_{test}^3 \dots \cup D_{test}^9$.

\subsubsection{Compare to Token-based Jailbreak}
We adopt Attack Success Rate (ASR) as the primary evaluation metric, computing it separately for each class on its corresponding $D_{train}$, $D_{test}$, and $D_{test}^{other}$. Tables \ref{white-box} and \ref{white-box2} present the jailbreaking performance of LLaMA2 and InternLM2.5 on the AdvBench-M dataset. Due to space limitations, the results for jailbreaking Qwen2.5 are provided in the supplementary material.

From Tables \ref{white-box} and \ref{white-box2}, we observe that our approach consistently outperforms all baselines across $D_{train}$, $D_{test}$ and $D_{test}^{other}$. On LLaMA2, our method demonstrates a substantial improvement over the second-best performer, GCG, with an average ASR increase of 2.9 percentage points on $D_{train}$ (85.0\% vs. 82.1\%) and a more significant 10.0 percentage point improvement on $D_{test}$ (84.7\% vs. 74.7\%). When evaluating on InternLM2.5, the performance advantage of our approach becomes even more pronounced, outperforming the second-best competitor, AdvAttack, by 21.6 percentage points on $D_{train}$ (82.8\% vs. 61.2\%) and by 26.6 percentage points on $D_{test}$ (84.6\% vs. 58.0\%). 

%From the results, we observe that our approach consistently outperforms competing methods across most harmful behavior categories on both models. For instance, on Llama-2, our method achieves 94\% ASR on $D_{test}$ for "Unlawful violence" compared to GCG's 83\% and AdvA's 60\%. Similarly, on InternLM2.5, our approach demonstrates superior performance across nearly all categories, particularly in "Cyber crimes" and "Fake info" where we achieve 96\% and 100\% ASR on $D_{test}$ respectively.

Notably, classes such as ``Suicide/Self-Harm'' and ``Hate'' remain particularly challenging to jailbreak. For instance, on LLaMA2, our method achieves only 44\% and 60\% ASR on $D_{test}$ for these classes respectively, and GCG reaches 38\% and 66\%. We attribute this difficulty to the fact that current models have undergone more intensive alignment efforts specifically for these sensitive classes. In contrast, certain classes demonstrate high vulnerability across most jailbreaking methods. For example, ``Financial Crimes,'' ``Property Crimes,'' and ``Cyber Crimes'' consistently yield high ASR across methods, with our approach achieving 98\%, 98\%, and 94\% respectively on $D_{test}$ for LLaMA2. This disparity reveals the inherent imbalance in current safety alignment mechanisms across different harm classes. These findings not only expose vulnerable areas in existing safety guardrails but also provide concrete guidance for more comprehensive and robust safety alignment in future LLM development.

%Notably, the class "Hate" proves challenging to jailbreak across both models. We speculate that recent LLMs have been specifically enhanced to defend against this category of harmful behaviors. Even with the introduction of visual modality, as our approach does, successfully jailbreaking this category remains difficult. This is likely because hate-related concepts are highly abstract, making it difficult to find suitable images to represent them. Thus, our visual module may \emph{not} have as significant an impact on the answer generation process of LLMs for this particular category.

%In contrast, for categories with strong visual representations such as "Weapons crimes," our approach significantly outperforms GCG, improving the ASR on Llama-2 from 66\% to 88\% on $D_{test}$. The cross-attention between visual input and textual queries becomes stronger for such concepts, thereby facilitating more successful jailbreaking.

%Regarding cross-class generalization capability, our approach still outperforms GCG. For example, for "Weapons crimes" on Llama-2, the ASR on $D_{test}^{other}$ is 66\% for GCG, while our approach achieves 78\% ASR.

%More importantly, the key advantage of our approach over discrete optimization-based methods lies in its efficiency. We compared the running time between our approach and GCG. Specifically, GCG takes 11.2 hours to find one txtJS for a single harmful class, whereas our approach requires only 0.37 hours—a substantial reduction in computational overhead.

\noindent\textbf{Efficiency.}  Our approach demonstrates markedly superior computational efficiency compared to token-based methods. Specifically, we compared the running time with GCG and found that GCG requires \textbf{11.2 hours} to generate a single \emph{JBtxt} for one harmful class, whereas our method completes the same task in just \textbf{0.37 hours}—approximately \textbf{3\% of GCG’s runtime}. This represents a substantial reduction in computational overhead.

\begin{table}[tpb]
\centering
%\hspace*{-0.5cm}
\resizebox{0.45\textwidth}{!}{
\setlength{\tabcolsep}{1.2pt} % Further reduced column spacing
\footnotesize % Reduced font size for the entire table
\begin{tabular}{
  >{\centering\arraybackslash}p{1.8cm}|
  >{\centering\arraybackslash}p{1cm}|
  >{\centering\arraybackslash}p{1.1cm}
  >{\centering\arraybackslash}p{1.1cm}
  >{\centering\arraybackslash}p{1.1cm}
  >{\centering\arraybackslash}p{1.1cm}
  >{\centering\arraybackslash}p{1.1cm}
  >{\centering\arraybackslash}p{1.1cm}
}
\toprule
\multirow{2}{*}{\textbf{Model}} & \multirow{2}{*}{\textbf{Dataset}} & 
\multicolumn{6}{c}{\textbf{Attack Success Rate (\%)}} \\
\cmidrule(lr){3-8}
& & \textbf{AutoP} & \textbf{AutoDAN} & \textbf{SoftP} & \textbf{GCG} & \textbf{AdvA} & \textbf{Ours} \\
\midrule
\multirow{3}{*}{LLaMA2} 
& ALL  & 0 & 0.5 & 16.0 & 32.0 & 67.5          & \textbf{72.0} \\
& TEST & 0 & 0.0 & 16.5 & 31.9 & 66.2          & \textbf{71.8} \\ 
& VAL  & 0 & 2.5 & 13.7 & 35.0 & \textbf{72.5} & \textbf{72.5} \\ 
\midrule
\multirow{3}{*}{InternLM2.5} 
& ALL  & 0 & 20.2 & 14.0  & 29.2  & 61.2 & \textbf{63.0} \\
& TEST & 0 & 18.7 & 13.4  & 27.5  & 59.3 & \textbf{64.0} \\ 
& VAL  & 0 & 26.2 & 16.2  & 36.2  & \textbf{68.7} & 59.0 \\ 
\bottomrule
\end{tabular}
}
\caption{White-box Jailbreak of LLaMA2 (via MiniGPT4-Jailbreak) and InternLM2.5 (via InternVL2.0-Jailbreak) on HarmBench dataset.}
\label{white-box-combined} 
\vspace{-1.0em}
\end{table}

%\subsubsection{Embedding Optimization for Jailbreaking}
\subsubsection{Compare to Embedding-based Jailbreak}
%For discrete optimization-based jailbreak, while it can directly find the jailbreaking suffix, it often suffers from efficiency issues due to the challenging nature of discrete optimization. In contrast, 
Compared to token-based jailbreak, embedding-based jailbreaking methods can leverage straightforward and efficient continuous optimization, since token embeddings are continuous variables. This technique is often referred to as ``soft prompting"\cite{lester2021power, qin2021learning, vu2021spot} in other literature and has demonstrated advantages in the Prompt Engineering domain. However, as pointed out in~\cite{zou2023universal}, this type of method is ineffective in LLM-jailbreak because the obtained embeddings (\emph{i.e.,} soft prompts) often have no corresponding token. In our experiments, we reached a similar conclusion to \cite{zou2023universal}, where the PEZ method completely failed to achieve any successful jailbreaks, \emph{i.e.,} $\text{ASR} \approx 0$ for all classes. Thus, the results of PEZ are omitted from Tables \ref{white-box} and \ref{white-box2}. The Soft Prompting (SoftP) method~\cite{qin2021learning} achieves slightly better performance than PEZ; however, its ASR remains lower than that of token-based methods, as shown in Tables \ref{white-box} and \ref{white-box2}.

We argue that this failure stems from the absence of constraints during the embedding optimization process. Without such constraints, the optimized embeddings tend to deviate significantly from the distribution of natural text embeddings, making it difficult to be converted back to text tokens. In contrast, our approach directly optimizes images, which are processed through the visual module. This ensures that the resulting visual embeddings inherently remain closer to the distribution of text embeddings. As a result, when projecting these visual embeddings into text tokens, the projection error is significantly reduced.

To validate our analysis, we quantitatively measure the error incurred when converting optimized embeddings back to tokens. For both the \emph{Soft Prompting} method and our approach, after converting \emph{JBemb} $e^k$ to \emph{JBtxt} via nearest neighbor search, we re-embed the obtained \emph{JBtxt} to produce $\hat{e}^k$. We then calculate the cosine similarity between $\hat{e}^k$ and $e^k$. A higher similarity score indicates a lower conversion error. In our experiments, the average similarity for soft prompting is $\text{Sim}_{sp}=0.17$, whereas our method achieves a substantially higher average similarity of $\text{Sim}_{ours}=0.63$. These results compellingly corroborate our analysis.

\subsubsection{Comparison on HarmBench}
Besides the AdvBench-M dataset, we also evaluate jailbreaking performance on the HarmBench dataset~\cite{HarmBench}. From Table~\ref{white-box-combined}, our approach significantly outperforms all baseline methods. Compared to AdvAttack, which shows strong performance on HarmBench, our method achieves a 4.5\% improvement on LLaMA2 (72.0\% vs. 67.5\% ASR on the entire dataset) and a 1.8\% improvement on InternLM2.5 (63.0\% vs. 61.2\% ASR), fully demonstrating the effectiveness of our approach.

\begin{table*}[ht]
%\vspace{-1.0em}
\centering
\setlength{\tabcolsep}{0.3em}
\resizebox{0.6\textwidth}{!}{
\begin{tabular}{c|c|c cc |c cc}
\toprule
\multicolumn{1}{c|}{\multirow{2}{*}{Initialization}}& 
\multirow{2}{*}{Clip-score}& 
\multicolumn{3}{c|}{MLLM} & 
\multicolumn{3}{c}{LLM} 
\\
& & $D_{train}$ & $D_{test}$ & $D_{test}^{others}$ & $D_{train}$ & $D_{test}$ & $D_{test}^{others}$
\\
\midrule
Random-based   & 0.1473 & 73.32 & 72.03 & 73.42 & 80.74 & 74.88 & 75.62 \\ 
Ranking-based  & 0.2083 & 75.18 & 76.75 & 75.44 & 82.22 & 77.07 & 75.94 \\
Ours           & 0.5773 & 83.70 & 80.13 & 76.61 & 87.40 & 85.03 & 78.99 \\
\bottomrule
\end{tabular}
\label{tb1}
}
\caption{Image-text Semantic Matching. We compare three initialization schemes.}
\vspace{-1.0em}
\end{table*}

\begin{table}[tpb]
\centering
\resizebox{0.45\textwidth}{!}{
%\hspace*{-1.0cm}
\setlength{\tabcolsep}{3.5pt}
\begin{tabular}{l|ccccc}
\toprule
\multirow{2}{*}{\textbf{Class}} & \multicolumn{5}{c}{\textbf{Black-box}} \\
 & Mistral & Gemma2 &  Deepseek-R1  &Grok-3 &DouBao1.5\\
\midrule
Violence  & 72.2 & 88.9 & 72.2  &  38.9  &5.6 \\
Financial    & 96.4 & 92.8 & 89.2  &  62.5  &39.2\\
Property     & 93.7 & 92.1 & 78.1  &  67.2  &31.2\\
Drug         & 86.7 & 80.0 & 86.7  &  26.7  &6.7 \\
Weapons      & 88.9 & 44.4 & 77.8  &  5.57  &11.1\\ 
Cyber        & 87.5 & 85.0 & 87.5  &  77.5  &35.0\\
Hate               & 66.7 & 53.3 & 40.0  &  60.0  &13.3\\
Self-Harm  & 22.2 & 61.1 & 16.7  &  66.7  &0.0 \\
Fake info          & 80.0 & 93.3 & 86.7  &  33.3  &40.0\\
\midrule
\textbf{Avg ASR}   & 77.1 & 76.7 & 70.5  &  48.7  &20.1\\
\bottomrule 
\end{tabular}
}
\caption{Black-box Jailbreak of different LLMs via InternVL2.0-Jailbreak.}
\label{Black-box} 
\vspace{-1.0em}
\end{table}

\subsection{Black-box Jailbreak} \label{black-boxsec}
For the black-box jailbreaking scenario, we generate the \emph{JBtxt} using a \emph{surrogate LLM} and transfer it to jailbreak the target LLM. In practice, we first construct an MLLM according to the surrogate LLM and then perform jailbreaking on this \emph{surrogate MLLM} to obtain the \emph{JBtxt}.

%Specifically, we conducted two experiments by selecting different surrogate LLMs. In the first experiment, we selected LLaMA-2 as the surrogate LLM. In practice, we employed MiniGPT-4 as the corresponding MLLM, as it contains a LLaMA-2 inside. We then performed jailbreaking on MiniGPT-4 to generate the \emph{JBtxt}. Finally, this \emph{JBtxt} was transferred to jailbreak target models, including Mistral-7B-v0.2, Gemma2-7B, Deepseek-R1, Grok-3, and DouBao1.5. The results are reported in Table~\ref{Black-box}. In the second experiment, we selected InternLM2.5 as the surrogate LLM. In practice, we employed InternVL2.0 as the corresponding MLLM, as it integrates a InternLM2.5 inside. We then performed jailbreaking on InternVL2.0 to generate the \emph{JBtxt}. The black-box jailbreaking results are shown in Table~\ref{Black-box2}.

In our experiments, we selected InternLM2.5 as the surrogate LLM. In practice, we employed InternVL2.0 as the corresponding MLLM, as it contains a InternLM2.5 inside. We then performed jailbreaking on InternVL2.0 to generate the \emph{JBtxt}. Finally, this \emph{JBtxt} was transferred to jailbreak target models, including Mistral-7B-v0.3, Gemma2-7B, Deepseek-R1, Grok-3, and DouBao1.5. The results are reported in Table~\ref{Black-box}. The experimental results demonstrate our approach has a strong black-box jailbreaking performance. We achieved remarkably high ASR of 77.1\%, 76.7\%, and 70.5\% against Mistral-v0.3, Gemma2, and Deepseek-R1, respectively. However, we observed comparatively moderate efficacy when targeting proprietary models, with ASR of 48.7\% on Grok-3 and 20.1\% on DouBao1.5, suggesting these proprietary models incorporate more robust defensive mechanisms against jailbreaking.

\subsection{Discussion}

\subsubsection{Image-text Semantic Matching}\label{ASR-preserving}
In our approach, MLLM-jailbreak is achieved by perturbing the \emph{JBinit} with a perturbation $\Delta$, \emph{i.e.,} $\text{JBimg} = \text{JBinit} + \Delta$. In this section, we illustrate that finding an appropriate \emph{JBinit} is crucial for both successful MLLM-jailbreak and subsequent LLM-jailbreak.
In our experiments, we compare three distinct schemes for finding \emph{JBInit}. The first scheme involves randomly sampling one image from all images in the AdvBench-M dataset, regardless of its harmful class. The second scheme restricts the sampling to images within the \emph{same} harmful class as the query, and further employs CLIP ViT-L/14 to identify the image that best aligns with the semantics of the corresponding class. We refer to this as the \emph{ranking-based} scheme. The third scheme is our proposed image-text semantic matching approach.

We compare the three initialization schemes in Table~\ref{tb1}. In terms of CLIP similarity score, the ranking-based scheme demonstrates a marginal improvement over the random sampling scheme. In contrast, our image-text matching scheme significantly outperforms both baselines, yielding substantially higher CLIP scores. This improvement stems from the fact that our method explicitly and \emph{directly} optimizes the \emph{JBInit} to maximize alignment with harmful queries in the CLIP embedding space.

In terms of jailbreaking ASR, we observe that selecting an appropriate \emph{JBInit} significantly boosts both MLLM-jailbreaking and LLM-jailbreaking ASR. Notably, as the CLIP similarity score increases, the ASR consistently improves.
%Take class1 as an example, although the ASR on $D_{train}$ is similar between the ranking scheme and our matching scheme, their ASR on $D_{test}$ for both MLLM- and LLM-jailbreak are simultaneously improved. 
%For instance, in class 4, despite having similar testing ASR for MLLM-jailbreaking, our matching scheme significantly outperforms the ranking scheme in terms of the testing ASR of LLM-jailbreaking.
%In comparison to the ranking scheme, which merely selects a semantically related image, our matching scheme enhances the matching score by modifying the InitJS. 
We argue that the underlying reason is that when the embedding of \emph{JBInit} is closely aligned with that of the harmful queries, \emph{JBInit} can significantly influence the LLM’s response generation—effectively turning refusal responses like ``Sorry, I cannot” into affirmative replies such as ``Sure, here is,” thereby enabling a successful jailbreak.

%From Table \ref{tb1}, we also observe that the LLM-jailbreaking ASR is better than the MLLM-jailbreaking ASR. This verifies that our double jailbreaking scheme can improve ASR in the white-box scenario.

\subsubsection{Cross-class Generalization}
In real-world jailbreak scenarios, it is desirable for the \emph{JBtxt} generated for a specific harmful class to generalize and effectively jailbreak other harmful classes.
%To evaluate this \emph{cross-class generalization} capability, we conduct an experiment where \emph{JBtxt} is generated based on a specific harmful class and subsequently tested on all remaining classes. For instance, after generating a \emph{JBtxt} using Class 1's training set $D_{train}^{1}$, we evaluate its ASR on the testing sets of the other classes: $D_{test}^{2}$, $D_{test}^{3}$, ..., $D_{test}^{9}$. This setup allows us to assess whether a \emph{JBtxt} optimized for one class can successfully jailbreak across different harm classes.
We conduct an experiment to evaluate the cross-class generalization capability of our approach. We find that certain classes, such as ``Financial Crimes", ``Property Crimes", and ``Cyber Crimes" exhibit strong generalizability. In contrast, classes like ``Hate" and ``Suicide and Self-Harm" prove challenging to generalize. The detailed results are provided in the supplementary material.

\section{Conclusion}
This paper introduces an efficient LLM-jailbreaking method by constructing a multimodal large language model and executing MLLM-jailbreak.
Unlike token-based jailbreaking techniques, our approach is notably efficient (approximately \textbf{3\%} of GCG’s runtime), as it exploits vulnerabilities within the visual module of the MLLM. Compared to embedding-based methods, we employ the visual module as a regularizer, ensuring that our jailbreaking embedding correspond to valid tokens. Furthermore, our approach exhibits superior cross-class generalization ability. Importantly, our \emph{indirect} jailbreaking scheme provides notable flexibility for black-box jailbreak, enabling successful jailbreaks of recent LLMs such as Mistral-v0.3, Gemma2, Deepseek-R1, Grok-3, and DouBao1.5.

\section{Acknowledgments}
This work was supported in part by Natural Science Basic Research Program of Shaanxi (Program No.2025JC-JCQN-075).

\bibliography{aaai2026}

\end{document}